# Automatic Design Method of Building Pipeline Layout Based on Deep Reinforcement Learning

Chen Yang, Zhe Zheng, Jia-Rui Lin*
Department of Civil Engineering, Tsinghua University, Beijing, China, 100084.
lin611@tsinghua.edu.cn (corresponding author)

**Abstract.** The layout design of pipelines is a critical task in the construction industry. Currently, pipeline layout is designed manually by engineers, which is time-consuming and laborious. Automating and streamlining this process can reduce the burden on engineers and save time. In this paper, we propose a method for generating three-dimensional layout of pipelines based on deep reinforcement learning (DRL). Firstly, we abstract the geometric features of space to establish a training environment and define reward functions based on three constraints: pipeline length, elbow, and installation distance. Next, we collect data through interactions between the agent and the environment and train the DRL model. Finally, we use the well-trained DRL model to automatically design a single pipeline. Our results demonstrate that DRL models can complete the pipeline layout task in space in a much shorter time than traditional algorithms while ensuring high-quality layout outcomes.

## 1. Introduction

In the construction industry, designing the layout of pipelines is a crucial task that involves transporting electricity, natural gas, and water from one point to another. This design is typically completed after determining the positions of the main components of the building. Currently, due to the relatively late adoption of automation and intelligence in the construction industry, pipeline layout design is still manually completed and highly relies on personal experience and regulatory requirements. Because unexpected issues may arise during the construction process, pipeline layout design often requires multiple rework designs, which consume significant amounts of manpower, materials, and financial resources (Singh, Deng and Cheng, 2018). Automated pipeline layout design can reduce the tediousness of this work and save time and money. Although pipeline layout design is a complex spatial geometry problem, it can be abstracted into a three-dimensional pathfinding and planning problem (Guirardello and Swaney, 2005), allowing for the automatic generation of pipelines. However, unlike general pathfinding problems, pipeline layout design requires not only meeting geometric constraints but also complying with numerous building code requirements. Consequently, finding the optimal solution for pipeline layout is a challenging task that belongs to NP-hard problems (Yates, Templeman and Boffey, 1984).

The most well-known and classic path-planning algorithm is the Dijkstra algorithm. However, it is often too time-consuming to be practical for large-scale application scenarios due to the NP-Hard nature of the problem. Path planning algorithms like Dijkstra and its derivatives experience rapid increases in search time with task complexity and constraints, making them impractical for complex tasks (Mukhlif and Saif, 2020). To address this, researchers have focused on heuristic algorithms such as A* algorithm (Singh and Cheng, 2020; Tsai et al., 2022) and fruit fly optimization algorithm (FOA) (Singh and Cheng, 2020) for pipeline layout design and optimization. These algorithms use heuristic functions to accelerate the search while balancing accuracy and efficiency to find better solutions faster. However, heuristic algorithms lack generalization ability and require recalculations for each new scenario or modification to building plans, which still consumes significant time.

In recent years, deep reinforcement learning (DRL) has matured and become increasingly popular as an alternative solution to heuristic algorithms. DRL is a machine learning method that combines deep learning (DL) and reinforcement learning (RL) (Arulkumaran et al., 2017). It involves sampling from the environment and training a deep neural network based on experience to provide the optimal solution for the agent. DRL can handle high-dimensional and continuous state and action spaces by using deep neural networks to approximate the value function or policy function in RL (Kaelbling, Littman and Moore, 1996). Compared to traditional algorithms, the most significant advantage of DRL is its ability to train a model with generalization capability through trial and error when tackling complex problems. The well-trained model can be applied directly in subsequent tasks without the need for further training, which saves a considerable amount of time. Additionally, DRL is more effective in solving complex problems. Although the training time of DRL increases when facing more complex tasks, the time spent on applying the well-trained model is roughly the same (Arulkumaran et al., 2017). Due to the advantages mentioned above, DRL has been extensively utilized in path-planning applications, such as drone path-planning problems (Huang et al., 2019; Bøhn et al., 2019). The primary objective of this paper is to investigate the potential applicability of DRL for 3D pipeline layout design. Additionally, we will analyse and compare its advantages and disadvantages with traditional methods.

## 2. Related Works

In the field of construction, pipeline layout design has gradually developed into a hot research area. Building information modeling (BIM), by combining mathematical modeling methods with collision detection technology, can intuitively analyse the position of complex node locations and detect collisions between intersecting nodes. The detected results can be manually corrected and have been widely applied in engineering. There is also research proposing a pipeline automatic correction method based on collision detection technology (Hsu et al., 2020). However, there is little work on pipeline layout design based directly on the geometric features of space and obstacles, and this field is still in its early stages.

Pipeline layout design can be abstracted into a three-dimensional path planning problem. Traditional algorithms for path planning include those based on graph theory searches, such as the Dijkstra algorithm and the A* algorithm; those based on random samplings, such as the rapidly-exploring random tree algorithm (RRT); and swarm intelligence algorithms, such as the genetic algorithm (GA) and particle swarm optimization (PSO) (Gasparetto et al., 2015; Aggarwal and Kumar, 2020). These algorithms have been successfully applied to solve shortest-path problems, including ship electromechanical pipeline layout design (Jiang et al., 2015), aviation engine pipeline layout design (Qu et al., 2016), and robot path planning (Ali et al., 2020). In the construction field, Singh and Cheng (2020) proposed a 3D multi-pipeline layout design automation method based on BIM and heuristic search methods. Tsai et al. (2022) proposed a field pipeline inspection and automatic coordination method based on augmented reality (AR) and the A* algorithm, which allows on-site personnel to compare newly added pipeline layout plans with existing pipelines. However, these algorithms still face the challenge of repeated computation and time waste in different scenarios.

RL enables agents to learn optimal strategies autonomously by maximizing cumulative rewards through trial-and-error interactions with the environment. RL has become a popular method in path planning. The traditional Q-learning algorithm (Watkins et al., 1992) stores all state-action pairs in a Q-table and searches for the optimal solution by referencing the Q-table. However, it is limited by its computational complexity when representing high-dimensional states. The introduction of Deep Q-Network (DQN) (Mnih et al., 2015) solves this problem. DQN is a value-based DRL algorithm that approximates the Q-value function using a deep neural network, allowing it to handle high-dimensional and discrete action spaces. After several years of development, more powerful DRL algorithms have been proposed. Proximal Policy Optimization (PPO) (Schulman et al., 2017) is one of the most widely applied algorithms. PPO is a policy gradient-based DRL algorithm that avoids performance degradation caused by overly large policy updates by limiting the magnitude of policy updates. The PPO algorithm is simple to implement and performs well in various tasks, making it one of the mainstream algorithms in the field of DRL. In the field of unmanned aerial vehicle (UAV) path planning, DRL has been used to address potential threats in complex and dynamic environments (Bøhn et al., 2019). This has implications for the practical application of DRL in the construction field.

## 3. Methodology

To address the above problem, this paper introduces the method of DRL, which aims to balance precision, generalization, and time. The process is illustrated in Figure 1 and comprises three modules: (1) establishing a training environment by abstracting building space models and analysing constraints, (2) training and optimizing the DRL model, and (3) evaluating the quality of the intelligently generated pipeline based on relevant indicators.

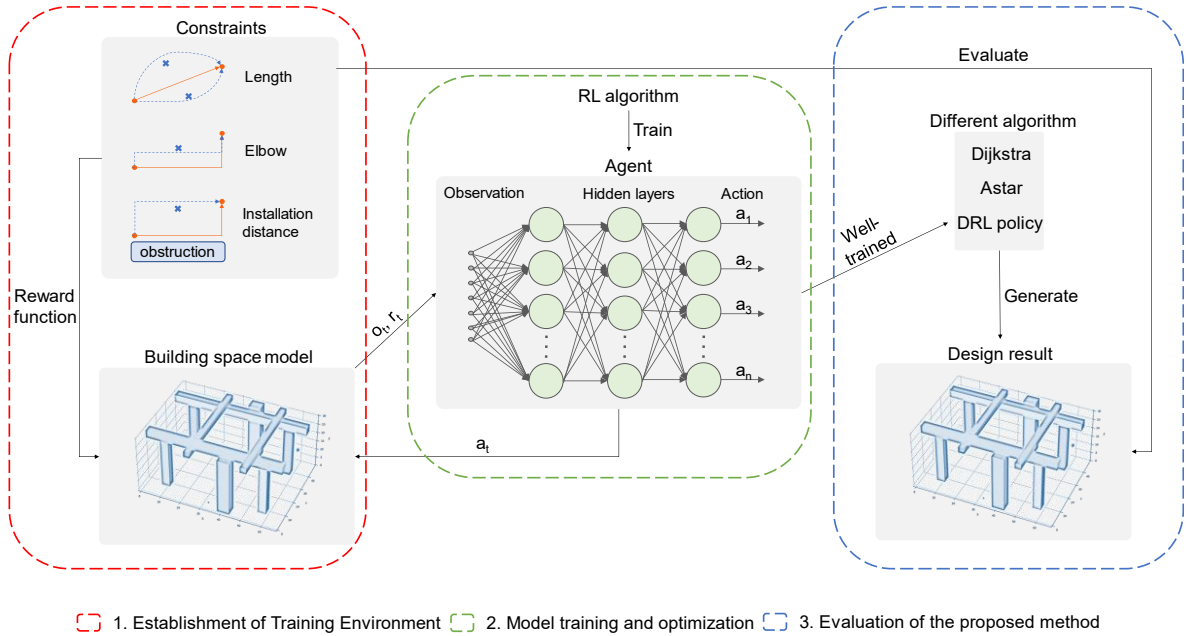

Figure 1: Methodology of the proposed method

### 3.1 Establishment of Training Environment

#### 3.1.1 Abstraction of Building Space Model

Real building environments are complex and contain numerous building elements that are too complex for model training and converging. Therefore, we analyse and retain the key elements in the space that affect pipeline layout to construct the training environment. The key elements retained include space size, positions and sizes of obstacles such as beams and columns, start and end positions of pipelines, and pipeline diameters. To further simplify the training environment, the space is rasterized into grids with a size of 10cm. On this basis, obstacles are simplified into cubic blocks parallel to the coordinate axis, and pipeline sections are simplified to 10cm x 10cm, occupying exactly one grid. This paper also defines a series of rules to ensure that the generated space complies with real building standards. For instance, the space dimensions must be between 5m-10m in length and width, with a height between 2.8m-4m. Additionally, the main beam size should be larger than the secondary beam size, and pipeline start and end points must be located on the wall.

Apart from the initialization function of the space, during training, this environment can also be reset before each episode, and the key elements inside the space will be randomly generated. This environment also has functions such as detecting collisions between the agent and obstacles or walls and determining whether the agent has reached the endpoint.

### 3.1.2 Geometric and Design Constraints Analysis

This paper mainly considers one geometric constraint and three design constraints for building pipeline layout design. The geometric constraint ensures that the pipeline does not collide with obstacles or walls and does not pass through the interior of obstacles or walls. The three design constraints include pipeline length, elbow, and installation distance. Short pipeline length is considered primarily for economic reasons, as it reduces the material cost of the pipeline and minimizes kinetic energy loss in the pipeline. The elbow angle should be as close to a right angle as possible to allow for the use of general elbow fittings, and the number of elbows should be minimized to further reduce kinetic energy loss. Installation distance refers to the pipeline's proximity to walls or obstacles, which should be as short as possible for easy installation and to reduce the length of fixing components. The reward function (described in Section 3.2.2) is utilized to train the agent and enforce these constraints.

## 3.2 Model Training and Optimization

### 3.2.1 Deep Reinforcement Learning Method

Compared with the classic supervised and unsupervised learning problems in machine learning, the biggest feature of RL is learning in interaction. The agent learns knowledge continuously according to the rewards or punishments obtained during the interaction with the environment and becomes more adapted to the environment. The learning paradigm of the agent is very similar to the process of human learning. The training process of DRL can be described by a quadruple <***Observation, Action, State transition probability, Reward***>. ***Observation*** is the environment state that the agent can perceive, namely the observation space. ***Action*** represents all the actions of the agent, namely the action space. ***State transition probability*** is the probability that the agent makes a certain action, namely the model and policy. ***Reward*** represents the reward or punishment corresponding to the action made in a certain state, namely the reward function (Kaelbling, Littman and Moore, 1996). The biggest difference between DRL and RL is that the model for action selection is a neural network. The core of applying DRL to pipeline design lies in designing observation space, action space, and reward function according to task needs, and selecting the appropriate algorithm and model. We present the details of the design in the next section.

### 3.2.2 Task Design Details

**Observation space design.** A common method for designing the observation space is to directly flatten the 3D matrix and input it into a deep neural network. However, in the training environment of this paper, the 3D building space model experiences random changes in size and obstacles. These changes can make it challenging to unify the input dimensions of the deep network. Additionally, due to the large size of the space, the flattened matrix dimension can be large, which may cause convergence issues. Therefore, this paper proposes a method that incorporates key parameters of the observation space to meet the task requirements. These parameters include relative coordinate, pipeline direction, start and end cube edge, angle, cross, and distance, as shown in Table 1. Finally, these parameters are flattened and input into a deep network for information fusion.

Table 1: Details of observation space design.

| Key parameters | Parameter definition | Legend |
|---|---|---|
| relative coordinate | Obtain the relative position of the start and stop points | \ |
| pipeline direction | Obtain pipeline direction | \ |
| start and end cube edge | Obtain the obstacle situation on the twelve edges of a cube diagonally connected to the starting and ending points | 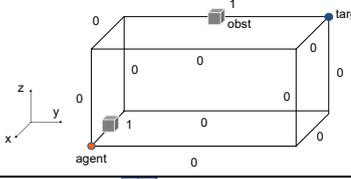 |
| angle | Obtain obstacles within a certain angle range | 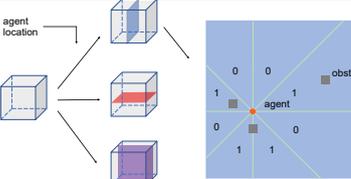 |
| cross | Obtain the obstacle situation on the cross in six directions | 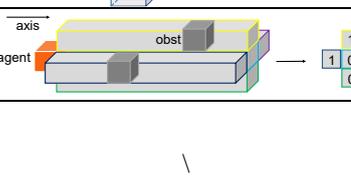 |
| distance | Obtain the distance between obstacles/walls in six directions | \ |

**Action space design.** After simplifying the pipeline, all elbow angles are 90 degrees. Therefore, the designed action space includes six actions: up, down, left, right, forward, and backward, representing movement in the positive and negative directions of the x, y, and z axes. To ensure model accuracy, the agent can only move one grid at a time and cannot move multiple grids simultaneously. To satisfy the geometric constraint that the pipeline cannot collide with obstacles or walls, an action masking mechanism is used to prohibit the agent from outputting actions that make it enter obstacles or walls directly. Previous studies have shown that action masking results in faster convergence and better performance than punishing actions with a certain size (Huang and Ontañón, 2022).

**Reward function design.** Based on the previous geometric and design constraint analysis, this section defines a series of reward functions. On one hand, for geometric constraints, the maximum reward of $R_{success} = +100$ is given for reaching the endpoint, which is a one-time reward for a successful layout. However, setting only a success reward means that the agent needs to go through more than a hundred steps to reach the endpoint and get a reward. The problem of "sparse rewards" can make training very difficult. To solve this problem, it is necessary to add denser rewards to guide the agent toward the endpoint. The agent receives a reward of $R_{closer} = +1$ when it moves closer to the endpoint and punishment of $R_{further} = -1$ when it moves further from the endpoint. On the other hand, for the three design constraints, three more rewards are designed. The agent receives a basic reward of $R_{base} = -0.5$ for every step it takes, preventing it from taking detours and reducing the length of the pipeline. The agent receives a punishment of $R_{elbow} = -5$ when it turns, reducing the number of elbows in the pipeline. The agent receives a punishment of $R_{install} = -w_{install} \times distance$, where $w_{install}$ is set to 0.15 and is proportional to the distance to the nearest obstacle. The weights of different rewards determine the final effect of the pipeline layout, and rewards that are too large or too small will increase the difficulty of training and may even prevent convergence.

**Algorithm and model selection.** The PPO algorithm was selected as the policy optimization algorithm for RL. Unlike standard policy gradient methods, which perform one gradient update per data sample, PPO can perform small batch updates for multiple epochs. The PPO algorithm belongs to the Actor-Critic method, which has two networks: one for generating policies (Actor) and the other for evaluating policies (Critic). Both networks are constantly updated, and this complementary training method is more effective than separate policy or value function networks. In this paper, the Actor and Critic share the underlying network, and the Actor outputs the probability of selecting each action while the Critic evaluates the current Actor.

### 3.3 Evaluation of the Proposed Method

Based on the constraints considered during training, the evaluation metrics for automatic pipeline design are defined from four aspects, length, elbow, installation distance, and layout time, respectively. Length refers to the total distance travelled by the agent from the starting point to the endpoint. Elbow refers to the number of times the agent turns. Installation distance is the average of the closest distance from the agent's position to a column, beam, or wall. Layout time is the total time it takes for the algorithm to complete the layout from start to finish. The values of these four indicators are all better when they are smaller. However, the three design constraint indicators are mutually restrictive and cannot be reduced simultaneously, so it is necessary to make trade-offs before evaluation.

## 4. Results and Discussion

### 4.1 Training Parameter Settings

The experiment was conducted using 1 Intel(R) Xeon(R) Silver 4215R 32-core CPU processor for parallel sampling and 1 NVIDIA GeForce RTX 2080 Ti GPU for accelerating training. Meanwhile, a parallel environment was created on 28 CPU cores to start sampling, with 28 agents collecting 8192 data samples each and feeding them into the GPU for training. The data was split into 8 batches, and the neural network was updated with 1024 samples per batch. The training process took approximately 2 hours to converge. In DRL, it is not suitable to use networks that are too large or too deep because the network needs to be frequently used for sampling and updating. Thus, a network with a depth of 4 layers and a hidden layer dimension of 512 is adopted.

### 4.2 Training Process

In training, adjusting the observation space and reward function based on an understanding of the task is more important than adjusting the hyperparameters related to DRL. On the one hand, the reward function needs to be designed based on the requirements of the task to ensure that the relative sizes of different rewards are appropriate, avoiding overly large or small rewards that may cause abnormal behaviour in the agents. On the other hand, it is necessary to filter out all relevant information required to meet the task requirements and pass it in a suitable unified form to the observation space. The design of an excellent observation space or reward function often plays a decisive role in the performance after convergence. Incomplete information or imbalanced rewards can lead to experimental failure. The following section presents a performance comparison of the optimal settings and some other settings during the training process.

### 4.2.1 Influence of Observation Space

The angle information in the observation space plays an important role in reducing the number of elbows and installation distances. By obtaining the current relative angular position of obstacles, agents can anticipate events that will occur within a certain number of steps in the future, thereby avoiding the selection of paths with more elbows or farther from the wall. If this information is deleted, the agent's level of confusion will increase, and it will be unable to learn the correct behaviour, as shown by the blue curve in Figure 2 (a). This is because the agent has been punished for some actions but does not know the reason for the punishment.

### 4.2.2 Influence of Reward Function

The weight distribution of different rewards in the reward function directly determines the effectiveness of the final pipeline layout. Therefore, the reward values need to be carefully adjusted. In this paper, the pipeline layout of the converged model was visualized under different reward functions, and the most realistic reward function was selected manually. Taking the elbow penalty in the reward function as an example, in the optimal setting, $R_{elbow} = -5$. When $R_{elbow} = -20$ is modified, the punishment value is too large, and the agent becomes timid, with a decreased desire to explore the endpoint and worse performance, as shown by the blue curve in Figure 2 (b).

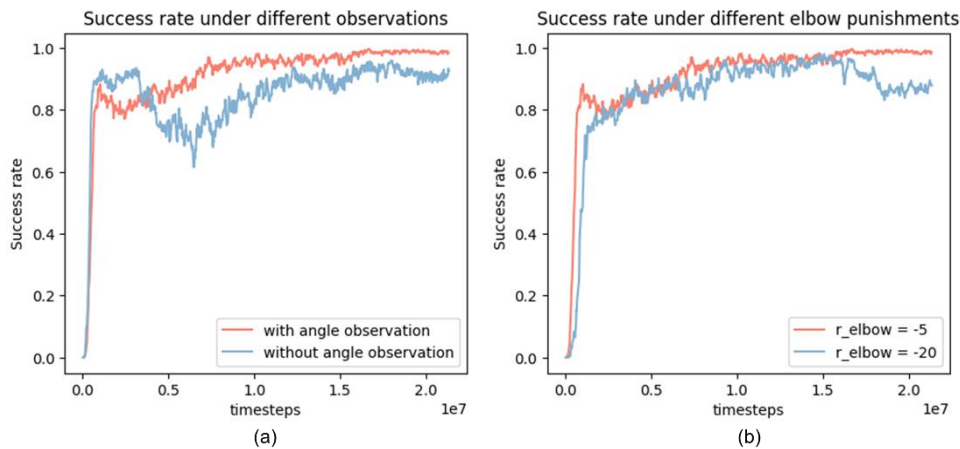

Figure 2: The Influence on Layout Success Rate of Different Observation Spaces and Reward Functions (Converged at 2e7 Timesteps)

### 4.3 Comparison with Classic Methods

To evaluate the performance of the models trained by DRL, two classic algorithms, Dijkstra and A* with the same constraints, are selected for comparison. Dijkstra algorithm is slower but more accurate and better equipped to find optimal solutions, providing an approximate optimal solution for design constraints. A* algorithm, on the other hand, is faster but may not always find the optimal solution, producing more balanced results. Using the same random seed, we added three constraints one by one (constraint 1: pipeline length, constraint 2: elbow, constraint 3: installation distance) to perform 100 layout designs.

The comparison results, as shown in Table 2, showed that for three or fewer constraints, the performance of the DRL had comparable capabilities compared to that of Dijkstra and A*. It is worth noting that as the number of constraints increased, the computing time of Dijkstra and A* algorithms increased sharply, particularly after the addition of the third constraint. This suggests that these two algorithms face time complexity challenges in more complex tasks. The DRL algorithm's running time remains roughly the same under different constraints and complexities, primarily related to the length of the pipeline path. For DRL, approximately 45% of the time is used to obtain observations, while the remaining time is utilized for model calculation, without the need for searches. This implies that the time efficiency of DRL will be much greater than that of other algorithms in more complex tasks.

Table 2: Comparison of model performance.

| Algorithm | Constraint | Length | Elbow | Installation distance | Time(s) |
|---|---|---|---|---|---|
| Dijkstra | 1 | **81.15** | **3.25** | 3.70 | 1.642 |
|  | 1, 2 | **81.15** | **2.32** | 4.87 | 1.928 |
|  | 1,2,3 | **83.83** | 4.02 | 1.44 | 11.544 |
| A* | 1 | **81.15** | 3.31 | 3.70 | **0.083** |
|  | 1, 2 | **81.15** | **2.32** | 4.87 | **0.114** |
|  | 1,2,3 | 83.99 | **3.99** | **1.42** | 1.603 |
| DRL | 1 | 83.63 | 9.82 | **3.58** | 0.189 |
|  | 1, 2 | 81.47 | 2.77 | **4.38** | 0.189 |
|  | 1,2,3 | 89.73 | 4.30 | 1.51 | **0.242** |

Figure 3 displays the results of pipeline layout designs by DRL, Dijkstra, and A* algorithms in the same typical scene. Among them, the pipeline layout results of Dijkstra and A* are the same. Both layout results exist in reality, and typically, one is selected based on the length and type of the pipeline.

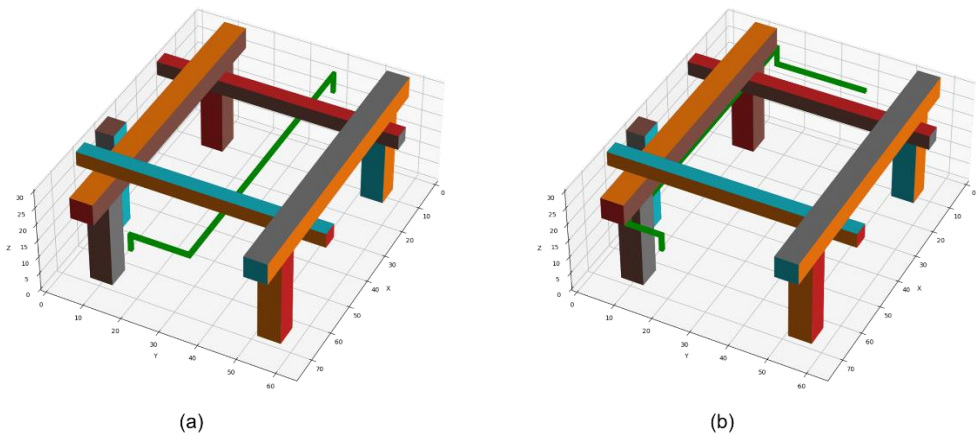

Figure 3: Pipeline Layout Using Different Algorithms in the Same Scenario
(a) DRL, (b) Dijkstra/A*

## 5. Conclusion and future works

This paper proposes an automatic pipeline layout design method based on DRL. The method consists of three main parts: building a training environment, model training and optimization, and evaluating the intelligent generation of pipeline quality. In the current experimental results, DRL has shown comparable capabilities to traditional methods in simple scenarios, but exhibits good potential in handling complex scenarios. The time taken by DRL to generate pipelines is not significantly impacted by the complexity of constraints, whereas other algorithms show a geometric growth trend. In addition, the current convergence performance may not represent the upper limit of DRL methods. Due to the large number of hyperparameters in DRL and multiple choices in the observation space, action space, reward function, and optimization algorithm, it is not so easy to fine-tune all the parameters and there is still potential to further improve the performance. Lastly, the training time of DRL in our case is about 2 hours, which is acceptable and negligible for practical applications.

In the future, we plan to continue testing the performance of DRL under more complex constraint conditions. As the task difficulty increases, the advantages of DRL may be fully demonstrated. Additionally, multi-pipeline layout design is a research direction we intend to pursue. We aim to apply multi-agent RL to multi-pipeline layout design to improve the overall efficiency and quality of pipeline systems.